\documentclass[10pt,a4paper,twoside,bibliography=totocnumbered]{article}

\usepackage{xcolor}

\usepackage{graphicx}

\usepackage{amsmath}
\usepackage{amssymb}

\usepackage{geometry}

\begin{document}

\title{{\huge \textcolor{black}{A Stochastic--Geometrical Framework for Object Pose Estimation based on Mixture Models Avoiding the Correspondence Problem }\\}}

\author{Wolfgang Hoegele$^{1}$\vspace*{0.3cm}\\
{\normalsize $^{1}$ Munich University of Applied Sciences HM}\\{\normalsize Department of Computer Science and Mathematics}\\{\normalsize Lothstraße 64, 80335 München, Germany}\vspace*{0.2cm}\\
{\normalsize \texttt{wolfgang.hoegele@hm.edu}}\vspace*{0.2cm}}

\date{\today}

\maketitle
\thispagestyle{empty}

\vspace*{1cm}
\textit{The article has been published under peer review and can be cited as:}\medskip

Hoegele, W. A Stochastic-Geometrical Framework for Object Pose Estimation Based on Mixture Models Avoiding the Correspondence Problem.\\
Journal of Mathematical Imaging and Vision (2024). https://doi.org/10.1007/s10851-024-01200-2
\vspace*{0.5cm}

\section*{Abstract}

\textit{Background:} Pose estimation of rigid objects is a practical challenge in optical metrology and computer vision. This paper presents a novel stochastic--geometrical modeling framework for object pose estimation based on observing multiple feature points.

\textit{Methods:} This framework utilizes mixture models for feature point densities in object space and for interpreting real measurements. Advantages are the avoidance to resolve individual feature correspondences and to incorporate correct stochastic dependencies in multi--view applications. First, the general modeling framework is presented, second, a general algorithm for pose estimation is derived, and third, two example models (camera and lateration setup) are presented.

\textit{Results:} Numerical experiments show the effectiveness of this modeling and general algorithm by presenting four simulation scenarios for three observation systems, including the dependence on measurement resolution, object deformations and measurement noise. Probabilistic modeling utilizing mixture models shows the potential for accurate and robust pose estimations while avoiding the correspondence problem. \medskip

\textbf{Keywords:} stochastic modeling, computer vision, object pose estimation, rigid body, camera system, lateration system\medskip

\newpage
\thispagestyle{empty}
\tableofcontents

\section*{About the Author}

Dr. Högele is a Professor of Applied Mathematics at the Department of Computer Science and Mathematics at the Munich University of Applied Sciences HM, Germany. His research interests are mathematical modeling, simulation and analysis of complex systems in applied mathematics with specific applications to radiotherapy, imaging, optical metrology and signal processing.

\thispagestyle{empty}

\newpage
\section{Introduction}

Pose estimation is a widely applied task in computer vision since the 1980's by Fischler \textit{et al.}\cite{Fischler1981} and other disciplines that require, for example, the automatic tracking of objects or single-- or multiple--view projection based estimation, i.e. see Groschl\"uter\textit{et al.}\cite{Groschlueter2022}, He \textit{et al.}\cite{He2021} and Sahin \textit{et al.}\cite{Sahin2020} for recent detailed reviews of state--of--the--art techniques. The goal is always to find the position and orientation of a predefined 3D object, which is observed typically by cameras.  To achieve this algorithmically, many advanced \textit{ad--hoc} loss functions and dedicated algorithms are applied successfully, most recently with the use of convolutional neural networks (CNNs) (e.g. Kendall \textit{et al.}\cite{Kendall2015}, Bui \textit{et al.}\cite{Bui2020}, Li \textit{et al.}\cite{Li2018}, and Wang \textit{et al.}\cite{Wang2020}) or kernel--based algorithms (e.g. Haopeng \textit{et al.}\cite{Haopeng2014} and Nie \textit{et al.}\cite{Nie2019}). { The exemplary background of this work is the rigid body pose estimation in the industrial application of tracking of laser scanning devices in optical metrology for which typically model based approaches, as presented in this paper, are state--of--the--art. The challenge for such devices is to find a balance between high accuracy and robust rigid body pose estimation, maintainability/servicability, safety and time performance regarding application and calibration of these products.} Useful approaches to a more general perspective on object pose estimation are modeling frameworks, which have been presented with the purpose to subdivide the process of pose estimation for multiple objects in a scene into smaller, modular steps and provide a review of possible specific algorithms of classical image processing, feature segmentation, image and 3D model matching and pose optimization (cp. Collet \textit{et al.}\cite{Collet2011}) or different architectures of CNNs for this purpose (cp. Li \textit{et al.}\cite{Li2018}). As demonstrated recently by He \textit{et al.}\cite{He2021}, the major challenges of 6D pose estimation to this day are the robustness of algorithms relative to image distortions and the problem of how to satisfyingly deal with deformed objects.\medskip

Probabilistic modeling in soft computing (e.g. Dogan\cite{Dogan2016} and Dote \textit{et al.}\cite{Dote2001}) and uncertainty quantification (e.g. Sullivan\cite{Sullivan2015}), which developed in parallel especially since the 1990's, has increasingly gained attention and importance in the real application (e.g. see Abdar \textit{et al.}\cite{Abdar2021}). The main advantage of this modeling approach is to directly model system components and/or situations with partial or probabilistic knowledge in a mathematical rigorous fashion and deduce direct algorithmical approaches. The focus of this paper is to apply mixture model density functions as a stochastic key component into the modeling of pose estimation, since this enables to combine probabilistic knowledge of individual features in object and observation space as well as the ignorance of which individual features are observed. \medskip

The closest applications of such a stochastic/probabilistic viewpoint to pose estimation has been applied by Erkent \textit{et al.}\cite{Erkent2016}, Teney \textit{et al.}\cite{Teney2014} and Teney \textit{et al.}\cite{Teney2011} with similarities especially by interpreting the observational measurement as a mixture model utilizing Gaussian kernels. Unfortunately only a partial stochastic model was applied which in turn led to \textit{ad--hoc} mathematical arguments and results, e.g. utilizing products of cross correlations in multi--view applications without strict derivations. Further, also the concept of modeling predicted marginal object point Gaussian Mixture Models directly in measurement space for 6D pose estimation has been introduced in Brachmann \textit{et al.}\cite{Brachmann2016} in order to increase the robustness of estimation. Although these approaches show high performance in their presented applications and prove the utility of stochastic models, in this paper the point of view is taken that the full potential of stochastic modeling for pose estimation in a general observation scenario is not addressed. For example, it is avoided to apply a clear and straightforward argumentation by random variables (including their dependency structure especially for multi--view applications), the applied density function types are limited to specific classes and the presented setting is limited to camera observations only. It is also noted that the stochastic modeling of feature points in object space (again limited to specific density functions, such as Gaussian Mixture Models) is currently successfully applied in point set registration, e.g. in Chen \textit{et al.}\cite{Chen2022} and Deng \textit{et al.}\cite{Deng2018}. Preliminary work for a specific example for a strictly stochastic framework for tumor pose estimation was presented in the field of medical imaging for radiotherapy with the purpose of optimal patient positioning by the author (Hoegele \textit{et al.}\cite{Hoegele2012,Hoegele2013}). \medskip

The motivation of the approach in this paper results directly from the presented challenges/limitations of current object pose estimations based on the observation of point features, such as deformed objects and observational distortions / noise, as well as presenting the full potential of probabilistic modeling based on mixture models for this application. Specifically, the goals are:
\begin{itemize}
\item Providing a thoroughly probabilistic argumentation of pose estimation based on a general stochastic--geometrical modeling framework utilizing mixture models. This will allow to understand and compare different observation setups/scenarios for pose estimation in a unified framework.
\item Showing how this modeling approach directly leads to a general algorithm for pose estimation resulting in a direct optimization problem. This includes the algorithms for single-- and multiple--view observations. This transparently derives a directly applicable algorithm based on the modeling framework in a strict fashion.
\item Providing expressive examples of specific observation setups for the application of this modeling framework in order to demonstrate its generality as well as utilization to specific observation scenarios.
\item Demonstrating the effectiveness of the resulting algorithm for specific observation scenarios by exemplary numerical simulations which, in addition, illuminate the opportunities of this framework for the real application. 
\end{itemize}

A specific advantage of the presented framework utilizing mixture model probability densities is the thorough avoidance of the correspondence problem, i.e. the individual classification/labeling and/or matching of features in measurements, as well as the linking of corresponding individual features in between observations and the model of the observed object. This is achieved by directly working with mixture model random variables in the object space as well as in the measurement space. Certainly the correspondence problem is solved successfully in many applications by iterative estimations applying robust statistics, e.g. see Collet \textit{et al.}\cite{Collet2011}. The major difference to these approaches is that within this novel framework the explicit solution of the correspondence problem is not necessary for pose estimation.\medskip

The structure of the paper follows directly the presented goals:\smallskip

First, in section \ref{sec:GenStochGeom_MM} an inclusive mathematical framework for many different observation applications is provided from a consistently stochastic perspective. The main ingredients are \textit{constructing a general feature appearance density function of the observational measurement as a mixture model}, \textit{introducing the feature positions in object space as a mixture model random variable} and \textit{a versatile description of the geometrical relation between object space and measurement space} (such as projections). Mathematically, this leads typically to nonlinear and not invertible transformations of the feature random variable.\smallskip

Second, in section \ref{sec:GenStochGeom_Alg} the general algorithm based on the stochastic--geometrical model is derived which leads to a single optimization problem and which allows pose estimation for single-- or multiple--view observations. In this context, the practical evaluation of these transformed random variables is presented by Monte--Carlo integration. It has to be pointed out, that in this optimization the correspondence problem is resolved implicitly.\smallskip

Third, in section \ref{sec:StochGeom_Examples} expressive examples for modeling specific observation scenarios are provided, including a classical observation scenario by cameras (which are widely utilized in computer vision) and a multi--lateration example (as an example which illustrates the possible variety of applications). The purpose of this section is to demonstrate the flexibility of this modeling framework but also to illustrate that these different observation scenarios can be investigated in a unified fashion.\smallskip

Fourth, in section \ref{sec:results} numerical simulations are presented to {indicate} the effectiveness, accuracy and robustness of the resulting {general} algorithm, depending specifically on observational distortions (i.e. on observation noise) as well as the deformation of the observed object.

\section{Methods}

This framework consists of the stochastic and geometric modeling of individual components of the observation process of an object. In this section, the relevant stochastic components are discussed and explained, the mathematical argument of the estimation of the object pose is presented and, finally, two mathematical applications of the modeling framework are demonstrated.

\subsection{General Stochastic--Geometrical Model}
\label{sec:GenStochGeom_MM}
\subsubsection{Interpretation of Processed Observation Measurements as Mixture Models}

First, we are focusing on the observational measurement data with a stochastic perspective. Such a probabilistic perspective on measurements has become recently more popular for object detection and pose estimation under the name \textit{probabilistic model of appearance} (see e.g. Erkent \textit{et al.}\cite{Erkent2016} and Teney \textit{et al.}\cite{Teney2014}). We denote $\vec{y}\in \mathbb{R}^{m}$ the coordinate in observation space, which means that the observation measurement is embedded in $\mathbb{R}^m$. Further, we describe $g_0(\vec{y})$ as the individual feature pattern representation after feature extraction in the observational measurement. We can call it also the \textit{sensing pattern} of the pose estimation. We normalize this function in order to interpret it as a mixture model probability density function (pdf) of the observation. As described, we are determining this pdf with a single observation which is different to the proposed \textit{probabilistic model of appearance} which utilizes a training image set.\smallskip

For example, if we observe point light sources placed on a rigid object with a camera, we can identify first all light sources by local thresholding and second, normalize the full image afterwards to get $g_0(\vec{y})$.\smallskip

Since we assume that the measurement and/or the feature extraction is not perfect, there is an inherent measurement or \textit{feature extraction error density} $\varepsilon(\vec{y})$ which we can gain from experience with the specific measurement device or feature extraction algorithm. This leads to a more valid description of this mixture model by the convolution of the sensing pattern with this error density $g(\vec{y})=g_0(\vec{y})\ast \varepsilon(\vec{y})$.\smallskip

Mathematically, we formally assign a random variable $\vec{Y}\sim g(\vec{y})$ to this density function. The thorough meaning of this random variable is challenging, since multimodal mixture density models can be hard to interpret in practice. In total, this random variable represents the probabilistic feature appearance pattern in the measurement.

\subsubsection{Modeling the Observation Prediction from Object Space to Measurement Space}
\label{sec:ObjectToMeasurement}

\subsubsection*{Object Description by Feature Coordinate Mixture Models}

We assume that the object contains observable point features. We denote $\vec{z}_i\in\mathbb{R}^{k}$ as the feature coordinates ($i=1,..,N$), which means that the original object is embedded in $\mathbb{R}^k$. Further, we assume that $\vec{z}_i$ is in general not known exactly and can be modeled individually by random variables $\vec{Z}_i\sim q_i(\vec{z})$ with corresponding probability density functions $q_i(\vec{z})$ representing the positioning probability. In order to present the general argumentation, we introduce the mixture model density
\begin{align*}
q(\vec{z}) := \frac{1}{N}\sum\limits_{i=1}^{N} q_i(\vec{z})\;,
\end{align*}
(without index) and assign the general random variable $\vec{Z}\sim q(\vec{z})$ to this mixture model in object space. The introduction of this mixture model feature density is one of the main strengths of this model, avoiding the correspondence problem.\smallskip

In general, it is not necessary to work with individual feature points $\vec{z}_i$ ($i=1,..,N$) to construct such a mixture model, but for the simplicity of the argumentation this is used throughout the paper. For example, also line segments (in order to approximate object edges) or other structures representing observable parts of the object and their uncertainties can be utilized in the construction of this mixture model. For example, other types of features could be surface structures of objects or even textures on an object surface. The crucial part will be that these features can be extracted well in the observations. For the practical implementation, the density function $q(\vec{z})$ needs to be modeled with high density areas at the location of these features.

\subsubsection*{Modeling the Geometric Observation Function}

The information about the geometry of observation is modeled by a function $\vec{p}(\cdot;\vec{\beta},\vec{\gamma}): \mathbb{R}^k\mapsto\mathbb{R}^m$. The observation obviously depends on the coordinate $\vec{z}\in\mathbb{R}^k$ and on observation specific parameters. We divide these parameters into two subsets:
\begin{itemize}
\item $\vec{\beta}\in\mathbb{R}^b$, the parameters describing the state of the object which we want to determine by the observation and typically represent the pose of the observed object (e.g. position, orientation, etc.). E.g. we could assume the $6$D pose transformation of a $3$D object to be described by $\vec{p}_1(\cdot;\vec{\beta}):\mathbb{R}^3\mapsto\mathbb{R}^3$, with $\vec{\beta}\in\mathbb{R}^6$.
\item $\vec{\gamma}\in\mathbb{R}^c$, the parameters of the parameterized setup of the observational system, which we know in advance (e.g. by calibration). E.g. we could assume the observational transformation of a $3$D object by $2$D projections into a image plane to be described by $\vec{p}_2(\cdot;\vec{\gamma}):\mathbb{R}^3\mapsto\mathbb{R}^2$.
\end{itemize}
In total, we can then describe the resulting observation including the pose transformation by $\vec{p}(\vec{z};\vec{\beta},\vec{\gamma}) := \vec{p}_2(\vec{p}_1(\vec{z};\vec{\beta});\vec{\gamma})$. It has to be noted that this function $\vec{p}$ is in general nonlinear and not invertible since typically $m<k$.\smallskip

An important class of transformations for the object transformation $\vec{p}_1$ are affine transformation in $\mathbb{R}^k$ with orthogonal matrices (length and angle conservation) plus a translation. This is the most important family of transformations when estimating the pose of rigid bodies since it includes to classic $6$D pose estimation of objects.

As an example, this can be modeled as a \textit{rigid affine transformation}: $\vec{p}_1(\vec{z}; (V,\vec{w}) )=V^T\cdot\left(\vec{z}+\vec{w}\right)$ with $V^T\in\mathbb{R}^{m\times k}$ (i.e. $V\in\mathbb{R}^{k\times m}$) and $V$ consisting of a $m$--dimensional orthonormal base in $\mathbb{R}^k$ and $\vec{w}\in\mathbb{R}^k$. The parameters $\vec{\beta}$ we want to ultimately determine by estimation are the entries of $V$ and $\vec{w}$. One way to look at it is the following: If we think of all possible linear transformations $A$ und represent them by the \textit{compact SVD}, i.e. $A=U\,\Sigma\,V^T$, the described restriction of the transformation matrix corresponds to $U=I_{k\times k}$ as the identity matrix and $\Sigma\in\mathbb{R}^{k\times m}$ consists of $1$s on the diagonal.

\subsubsection*{The Predicted Observation as a Mixture Model in Measurement Space}

An important conceptual step is to combine the stochastic description of the features in object space with the geometric transformation. In general terms we denote $\vec{p}\left(\vec{Z};\vec{\beta},\vec{\gamma}\right)$ as the random variable transformation of the mixture model depending on the object features $\vec{Z}$. It is obvious that this transformation of the mixture model leads in general also to mixture model predictions in measurements space.\smallskip

Calculating the probability density of $\vec{p}\left(\vec{Z};\vec{\beta},\vec{\gamma}\right)$ analytically is typically difficult or even impossible, since $\vec{p}$ is in general \textit{nonlinear} and \textit{not invertible}. It has to be noted that numerical approximations of challenging pdf transformations can always be performed utilizing \textit{Monte--Carlo}--Methods, as presented in the following section.

\subsection{General Algorithm based on the Stochastic--Geometrical Model}
\label{sec:GenStochGeom_Alg}

\subsubsection{Modeling the Object Pose Estimation with a Single Observation}

The matching of observation and prediction utilizing these mixture models is presented in a straightforward argumentation. We recall, $\vec{Y}\sim g(\vec{y})$ as the true measurement random variable and $\vec{p}\left(\vec{Z};\vec{\beta},\vec{\gamma}\right)$ as the independently constructed observation random variable depending on the parameters $\vec{\beta},\vec{\gamma}$. An intuitive argument is that we want both random variables to match as well as possible, which we simply put as
\begin{align*}
\vec{Y}\stackrel{!}{=} \vec{p}\left(\vec{Z};\vec{\beta},\vec{\gamma}\right)\;\Leftrightarrow\; \vec{Y}-\vec{p}\left(\vec{Z};\vec{\beta},\vec{\gamma}\right) \stackrel{!}{=} \vec{0}\;.
\end{align*}
We interpret this straightforwardly in the following way: The probability density function of the difference random variable $\vec{Y}-\vec{p}\left(\vec{Z};\vec{\beta},\vec{\gamma}\right)$ must have the highest possible value at $\vec{0}$.
The difference random variable $\vec{Y}-\vec{p}\left(\vec{Z};\vec{\beta},\vec{\gamma}\right)$ at position $\vec{0}$ is distributed by
\begin{align*}
f_{\vec{Y}-\vec{p}\left(\vec{Z};\vec{\beta},\vec{\gamma}\right)}(\vec{0}) &= \int_{\mathbb{R}^k} f_{\vec{Y}-\vec{p}\left(\vec{z};\vec{\beta},\vec{\gamma}\right)}(\vec{0}) \cdot q(\vec{z}) \;\text{d}\vec{z}\;,
\end{align*}
utilizing the law of total probability for densities. By inserting the mixture model for $\vec{Z}$ we get
\begin{align*}
&= \int_{\mathbb{R}^k} f_{\vec{Y}-\vec{p}\left(\vec{z};\vec{\beta},\vec{\gamma}\right)}(\vec{0}) \, \left(\frac{1}{N}\sum\limits_{i=1}^{N} q_i(\vec{z})\right) \;\text{d}\vec{z} \\
&= \frac{1}{N}\sum\limits_{i=1}^{N} \int_{\mathbb{R}^k} f_{\vec{Y}-\vec{p}\left(\vec{z};\vec{\beta},\vec{\gamma}\right)}(\vec{0})\cdot q_i(\vec{z})\;\text{d}\vec{z}\;.
\end{align*}
By realizing that $\vec{Y}-\vec{p}\left(\vec{z};\vec{\beta},\vec{\gamma}\right)$ is simply the shifted measurement random variable, we arrive at
\begin{align*}
&= \frac{1}{N}\sum\limits_{i=1}^{N} \int_{\mathbb{R}^k} g\left(\vec{p}\left(\vec{z};\vec{\beta},\vec{\gamma}\right)\right)\cdot q_i(\vec{z})\;\text{d}\vec{z}\;.
\end{align*}
Following the argument of the highest possible value at $\vec{0}$, the matching leads to the maximization \begin{align}
\text{arg max}_{\,\vec{\beta}}\; f_{\vec{Y}-\vec{p}\left(\vec{Z};\vec{\beta},\vec{\gamma}\right)}(\vec{0})&=\text{arg max}_{\,\vec{\beta}}\; \frac{1}{N}\sum\limits_{i=1}^{N} \int_{\mathbb{R}^k} g\left(\vec{p}\left(\vec{z};\vec{\beta},\vec{\gamma}\right)\right)\cdot q_i(\vec{z})\;\text{d}\vec{z}\;.
\label{equ:singleobserv_objfunc}
\end{align}

\subsubsection*{General Implementation via Monte--Carlo}

This formula (\ref{equ:singleobserv_objfunc}) can be approximated directly by Monte--Carlo integration (see e.g. Erkent \textit{et al.}\cite{Erkent2016} and Teney \textit{et al.}\cite{Teney2014}). First, $R$ samples $\vec{z}_{i,r}$ ($r=1,..,R$) of each individual feature coordinate $q_i(\vec{z})$ are drawn, second, the transformation is applied $\vec{p}(\vec{z}_{i,r};\vec{\beta},\vec{\gamma})$ to each sample and, third, these values are utilized in the approximation of the integral via Monte--Carlo, resulting into:
\begin{align}
\text{arg max}_{\,\vec{\beta}}\; \frac{1}{N R}\sum\limits_{i=1}^{N} \sum\limits_{r=1}^{R}  g\left(\vec{p}\left(\vec{z}_{i,r};\vec{\beta},\vec{\gamma}\right)\right)\;.
\label{equ:singleobserv_objfunc_MC}
\end{align}
In this implementation one needs to perform many interpolations of the measurement density function $g$ in a fast fashion, for example, such as bilinear interpolation in a $2$D measurement space.

\subsubsection*{Interpretation}

A general stochastic-geometrical derivation is presented that allows object pose estimation without resolving the correspondence of individual features between measurement and model prediction. Of course, having not enough features in the measured density $g_0(\vec{y})$ or not enough features $N$ at all in the 3D object the pose estimation will not be able to find a unique or distinct global maximum of this objective function. The benefit of this argument is that these ambiguities of the application are directly transferred to non--unique maxima of the optimization problem. \smallskip

During optimization the density $\varepsilon(\vec{y})$, which was introduced as the error density regarding to feature extraction in the observation, can be utilized in a practical way. One can start optimization utilizing $\varepsilon(\vec{y})$ with an artificially higher standard deviation and successively reduce it during optimization until it is at a realistic range, leading to broader peaks in the objective function which might be beneficial for optimization. Then the meaning of this error density would shift to be a \textit{sensing density}, i.e. that we artificially blur the observation in the beginning of optimization for a rough alignment (corresponding then to broader and shallower local maxima of the objective function) and increasing accuracy utilizing sharper error densities (corresponding to narrower and deeper local maxima).\smallskip

It is a powerful extension of this framework interpreting the density functions of feature coordinates of the object $\vec{Z}_i\sim q_i(\vec{z})$ not only as the \textit{unknown position} of a truly rigid body but also of a deformed object. Of course, this stochastic formulation of object pose estimation is then still straightforwardly applied without change for this much more difficult task since it is only a change in the interpretation of these density functions. For example, when tracking large and/or heavy rigid objects it is a practical concern that they slightly deform depending on the $6$D pose since the gravitational force is acting on the whole body. This may lead to individual feature displacements depending on the $6$D pose, which (moderately) violates the rigid body assumption. In the presented approach it is practically possible to incorporate such feature displacements as positional uncertainties into the pose estimation by updating the feature density functions with the range of individual feature displacements in the direction of the gravitational force.\smallskip

Further, it is a legitimate point to question if the proposed maximization of $f_{\vec{Y}-\vec{p}\left(\vec{Z};\vec{\beta},\vec{\gamma}\right)}(\vec{0})$ is an \textit{ad hoc} assumption. First, it is stressed that this argument is the most direct and simple way of interpretation and, therefore, regarded as very general. Second, one can extend this, of course, in order to evaluate the concentration of the density around $\vec{0}$ by averaging
\begin{align*}
\text{arg max}_{\,\vec{\beta}}\; \int\limits_{\mathbb{R}^m} w(\vec{s})\cdot f_{\vec{Y}-\vec{p}\left(\vec{Z};\vec{\beta},\vec{\gamma}\right)}(\vec{s})\;\text{d}{\vec{s}}\;,
\end{align*}
with a non-negative weighting function $w:\mathbb{R}^m\mapsto\mathbb{R}$ concentrated around $\vec{0}$ which shapes such an averaging process. 

\subsubsection*{Validation Example}

As a theoretical validation example we are utilizing a camera system with exactly known feature coordinates in $3$D object space and an unspecified projection geometry. This means, we are observing a $3$D object with feature coordinates $\vec{z}_i\in\mathbb{R}^3$, ($i=1,..,N$), with $2$D camera observations/projections. In this validation example we assume to know the exact positions of the feature coordinates, i.e. $q_i(\vec{z})=\delta(\vec{z}-\vec{z}_i)$ applying the delta distribution as artificial density function. This leads to the mixture model density for $\vec{Z}$
\begin{align*}
q(\vec{z}) = \frac{1}{N} \sum\limits_{i=1}^N \delta(\vec{z}-\vec{z}_i)\;.
\end{align*}\smallskip

We utilize a general projection function in this example $\vec{p}(\vec{z};\vec{\beta},\vec{\gamma})$ without further specification. For example, it could be parallel beam or cone beam projection geometry for a pinhole camera.\smallskip

This leads to the following objective function for one projection (utilizing the \textit{sifting} property of the delta distribution), applying formula (\ref{equ:singleobserv_objfunc}):
\begin{align}
\text{arg max}_{\,\vec{\beta}}\; \frac{1}{N} \sum\limits_{i=1}^{N} g\left( \vec{p}\left(\vec{z}_i;\vec{\beta},\vec{\gamma}\right) \right)\;.
\label{equ:validexample}
\end{align}
This is a rather intuitive result, since this approach maximizes the sum of point samples of the measurement mixture model density that correspond to the projected positions $\vec{p}(\vec{z}_i;\vec{\beta},\vec{\gamma})$. Since we have only delta distributions, no numerical approximation of density functions via Monte--Carlo is necessary.\smallskip

The validation of the presented mathematical arguments gets obvious if we assume that the observed body is fully rigid (i.e. no deformations are present) and the observation of the features has no bias, i.e. the mixture model observation pdf $g(\vec{y})$ has its (local) maxima exactly at the positions given by the projection geometry of the known feature coordinates. Then equation (\ref{equ:validexample}) exactly samples and sums up the (local) maxima of $g(\vec{y})$, which in consequence corresponds to the global maximum of the objective function. This validates the algorithmic approach for a single observation.

\subsubsection{Extension to Multiple Observations with known Relative Geometry}

This object pose estimation can be easily extended to multiple observations if the relative geometry of the observations are fixed and known in advance, e.g. by a static multiple camera device. The purpose again, is to find the parameters $\vec{\beta}$ describing the status of the object relative to the observation geometry as in the previous derivations.\smallskip

We introduce the general notation for parameters $\vec{\gamma}_l$ describing the state of each individual observation device (e.g. position and orientation of the measurement device) in a global coordinate system for $l=1,..,L$ observation devices.\smallskip

Before we arrive at the main argument, we must clarify an important point: A common misconception is that the matching functions between observation and prediction for different observations of the same object are regarded as independent. In consequence, these matching functions of different observations $l=1,..,L$ are typically multiplied. It is pointed out that this is wrong, which can be realized by working directly with the difference random variables $\vec{Y}_l-\vec{p}\left(\vec{Z};\vec{\beta},\vec{\gamma}_l\right)$ ($l=1,..,L$). These difference random variables are not independent for different observations $l$, since the prediction part $\vec{p}\left(\vec{Z};\vec{\beta},\vec{\gamma}_l\right)$ depends on the commonly observed object features described by $\vec{Z}$. To illustrate the meaning of this argument, it is emphasized that the obvious multiplication of matching functions of each individual observation misses the correct dependency structure, e.g.
\begin{align*}
 f_{\bigcap\limits_{l=1}^L \vec{Y}_l-\vec{p}\left(\vec{Z};\vec{\beta},\vec{\gamma}_l\right)}(\vec{0}) \,\neq\, \prod\limits_{l=1}^L\, f_{ \vec{Y}_l-\vec{p}\left(\vec{Z};\vec{\beta},\vec{\gamma}_l\right)}(\vec{0})\;,
\end{align*}
with $f_{\bigcap\limits_{l=1}^L \vec{Y}_l-\vec{p}\left(\vec{Z};\vec{\beta},\vec{\gamma}_l\right)}$ being the joint probability distribution of the matching for all observations.

On the other side, the individual measurement random variables $\vec{Y}_l$ can safely be assumed to be independent, especially for electronic measurement devices which have no \textit{memory} in their sensors. \medskip

To avoid this misconception, we have to introduce the following argumentation if we consider the correct dependency structure of the random variables
\begin{align*}
f_{\bigcap\limits_{l=1}^L \vec{Y}_l-\vec{p}\left(\vec{Z};\vec{\beta},\vec{\gamma}_l\right)}(\vec{0}) &= \int_{\mathbb{R}^k} f_{\bigcap\limits_{l=1}^L \vec{Y}_l-\vec{p}\left(\vec{z};\vec{\beta},\vec{\gamma}_l\right)}(\vec{0}) \cdot q(\vec{z}) \;\text{d}\vec{z}\;,
\end{align*}
utilizing again the law of total probability for densities. By inserting the mixture model for $\vec{Z}$ and realizing that $\vec{Y}_l-\vec{p}\left(\vec{z};\vec{\beta},\vec{\gamma}_l\right)$ are independent for different $l$ (since the measurements themselves are independent) we get
\begin{align*}
&= \frac{1}{N}\sum\limits_{i=1}^{N} \int_{\mathbb{R}^k} \prod\limits_{l=1}^L\, f_{\vec{Y}_l-\vec{p}\left(\vec{z};\vec{\beta},\vec{\gamma}_l\right)}(\vec{0}) \cdot  q_i(\vec{z}) \;\text{d}\vec{z}
\end{align*}
\begin{align}
&= \frac{1}{N}\sum\limits_{i=1}^{N} \int_{\mathbb{R}^k} \prod\limits_{l=1}^L\, g_l\left(\vec{p}\left(\vec{z};\vec{\beta},\vec{\gamma}_l\right)\right)\cdot q_i(\vec{z})\;\text{d}\vec{z}\;.
\label{equ:multiobserv_objfunc}
\end{align}
Applying the Monte--Carlo approximation to the objective function, we finally arrive at
\begin{align}
\text{arg max}_{\,\vec{\beta}}\; \frac{1}{N R} \sum\limits_{i=1}^{N} \sum\limits_{r=1}^{R} \prod\limits_{l=1}^L\, g_l\left(\vec{p}\left(\vec{z}_{i,r};\vec{\beta},\vec{\gamma}_l\right)\right) \;,
\label{equ:multiobserv_objfunc_MC}
\end{align}
again with $\vec{z}_{i,r}$ ($r=1,..,R$) being random samples of the pdf $q_i(\vec{z})$ ($i=1,..,N$). \medskip

For comparison of the formula, the result of the neglection of the dependency structure by multiplication of the matching function of different observations leads to the Monte--Carlo approximation
\begin{align*}
\text{arg max}_{\,\vec{\beta}}\; \frac{1}{(N R)^L}\prod\limits_{l=1}^L \left(\sum\limits_{i=1}^{N} \sum\limits_{r=1}^{R}  g_l\left(\vec{p}\left(\vec{z}_{i,r};\vec{\beta},\vec{\gamma}_l\right)\right) \right)\;,
\end{align*}
with $\vec{z}_{i,r}$ ($r=1,..,R$) being random samples of the pdf $q_i(\vec{z})$ ($i=1,..,N$).\smallskip

The difference constitutes itself by the different position of the product. It is probable that utilizing a wrong assumption, such as the independency of matching functions in a multi--view application, leads to inferior results.

\subsubsection*{Calibration Of Relative Geometry of Observation Devices}

It will now be demonstrated that this formula (\ref{equ:multiobserv_objfunc}) can also be utilized to perform the calibration of an observation system consisting of multiple observations. In this case we are inserting a suitable object with known or defined parameters  $\vec{\beta}$ (e.g. orientation and position) in the observational device (or practically also often utilized is the mounting of the observation devices in a larger calibration system) and are maximizing with respect to the observation geometry parameters $\vec{\gamma}_l$
\begin{align*}
\text{arg max}_{\,\vec{\gamma}_{1},\dots,\vec{\gamma}_{L}\,}\; \frac{1}{N}\sum\limits_{i=1}^{N} \int_{\mathbb{R}^k} \prod\limits_{l=1}^L\, g_l\left(\vec{p}\left(\vec{z};\vec{\beta},\vec{\gamma}_l\right)\right)\cdot q_i(\vec{z})\;\text{d}\vec{z}\;.
\end{align*}
It has to be pointed out that in calibration one typically is not interested in the absolute parameter optimization of each observation geometry $\vec{\gamma}_l$, since this leads to ambiguities, for example, if the whole detection system including all observations is geometrically transformed as a rigid body. In consequence, one is particularly interested in the relative difference of the observation geometries, which, for example, can be easily realized by setting some parameters of $\vec{\gamma}_l$ to fixed values (in order to prevent such global device transformations during calibration) while the others are optimized during calibration. This point is further illustrated in the application examples in subsection \ref{sec:StochGeom_Examples}.

\subsubsection*{Extension of the Validation Example}

Following the argumentation of the validation example for the single observation, the direct extension to $L$ projections leads to the extension of formula (\ref{equ:validexample}):
\begin{align*}
\text{arg max}_{\,\vec{\beta}}\; \frac{1}{N} \sum\limits_{i=1}^{N} \prod\limits_{l=1}^{L}\,  g_l\left( \vec{p}\left(\vec{z}_i;\vec{\beta},\vec{\gamma}_{l}\right) \right)\;.
\end{align*}

Again, the validation gets obvious if we assume the same rigid body setup and no observation bias is present as for the single observation case. Then the true (local) maxima of the mixture model observation pdfs $g_l(\vec{y})$ are first multiplied for each single feature coordinate $\vec{z}_i$ with respect to the projections, and, second, are summed up for all features. This again corresponds to the global maximum of this objective function and, in consequence, validates the algorithm for multiple observations.

\subsection{Examples for the Application of the Modeling Framework}
\label{sec:StochGeom_Examples}

The power of the mathematical framework will be illustrated by presenting relevant applications. In consequence, two example applications for object pose estimation are demonstrated within this framework. It has to be pointed out that the general framework, of course, is not limited to these applications, but they exemplify how this framework can be utilized practically. 

\subsubsection{Short Tutorial for Applying the Stochastic--Geometrical Modeling}

In order to allow a straightforward application of the modeling framework, a brief summary in a tutorial style is presented:
\begin{itemize}
\item[a)] Extract the probabilistic mixture model observation density functions $g_l(\vec{y})$ for each observation $l=1,..,L$, which shows the observed features. This might include signal or image processing for feature extraction in real measurements as well as applying the error density $\varepsilon(\vec{y})$.
\item[b)] Probabilistically model the features in object space by a mixture model object density function $q(\vec{z})$. For example, if the observable object consists of several point features, model the knowledge level about the point coordinates by corresponding probability density functions.
\item[c)] Model the deterministic geometric observation function $\vec{p}(\cdot;\vec{\beta},\vec{\gamma}_l)$, especially, which parameters one wants to determine by the observation $\vec{\beta}$ (such as a $6$D pose) and which are assumed/calibrated beforehand by the observation setup $\vec{\gamma}_l$ for each observation $l=1,..,L$.
\item[d)] Formulate and perform the maximization of equation (\ref{equ:multiobserv_objfunc}), for example, utilizing Monte--Carlo integration as in equation (\ref{equ:multiobserv_objfunc_MC}) in order to estimate $\vec{\beta}$.
\end{itemize}

In the following two subsections, specific examples for such stochastic--geometrical models are presented. Since this should only demonstrate the modeling part, point a) of the list is not executed and point d), in consequence, is not performed but only formulated.

\subsubsection{Camera System with Feature Coordinate Densities and Parallel Beam Projection Geometry}
\label{sec:cameraexample}

In this example application, we are observing a $3$D object with feature coordinates $\vec{z}_i\in\mathbb{R}^3$, ($i=1,..,N$), with $2$D camera observations/projections. We assume a low knowledge level regarding the feature coordinates, specifically, that they are symmetrically normal distributed, i.e. $q_i(\vec{z})=\mathcal{N}(\vec{z}_i,\text{diag}(\sigma_{i,1}^2,\sigma_{i,2}^2,\sigma_{i,3}^2))$.\smallskip

\begin{figure}[htbp]
\centering
\includegraphics[width=8cm]{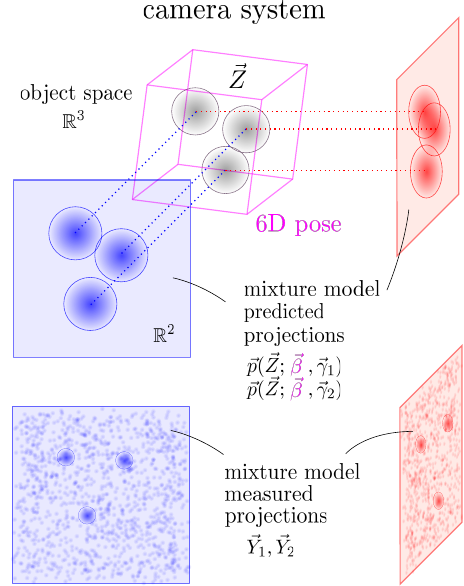}
\caption{Schematic illustration of a camera system observation geometry and their connections to the mixture model approach.}
\label{OverviewPlot_A}
\end{figure}

Further, we assume a specific projection geometry: Parallel beam projection from 3D-object to 2D-camera while the 3D-object can be shifted diagonally and rotated only around the $z_3$-axis (this limited transformation will be used in the results subsection \ref{sec:res_vis} for visualization purposes):
\begin{align*}
\vec{p}_1(\vec{z};(\varphi,w)) &:= \left(\begin{array}{c}
\cos(\varphi)\cdot (z_{1}+w) + \sin(\varphi)\cdot (z_{2}+w)\\
-\sin(\varphi)\cdot (z_{1}+w) + \cos(\varphi)\cdot (z_{2}+w)\\
z_{3}+w
\end{array}\right)\;,\\
\vec{p}_2(\vec{p}_1(\vec{z};(\varphi,w));\gamma) &:= \left(\begin{array}{ccc}
\cos(\gamma) & \sin(\gamma) & 0\\
0 & 0 & 1
\end{array}\right)\cdot\vec{p}_1(\vec{z};(\varphi,w))\;.
\end{align*}
The parameter $\gamma$ describes in this case how the cameras are rotated around the object. The overall measurement situation is illustrated in figure \ref{OverviewPlot_A}.\smallskip

We can use the Monte--Carlo approximation, which leads to the specific application of formula (\ref{equ:singleobserv_objfunc_MC}):
\begin{align*}
\text{arg max}_{\,(\varphi,w)}\; \frac{1}{N\, R} \sum\limits_{i=1}^{N} \sum\limits_{r=1}^R g\left(\left(\begin{array}{ccc}
\cos(\gamma) & \sin(\gamma) & 0\\
0 & 0 & 1
\end{array}\right)\cdot\vec{p}_1(\vec{z}_{i,r};(\varphi,w))\right)\;,
\end{align*}
with $\vec{z}_{i,r}$ ($r=1,..,R$) drawn from the described normal distributions $\mathcal{N}(\vec{z}_i,\text{diag}(\sigma_{i,1}^2,\sigma_{i,2}^2,\sigma_{i,3}^2))$.

For the extension to multiple projections, we present the formula for pose estimation with two orthogonal projections, which means that we have $\gamma_1$ as initial angle of the first observation and $\gamma_2:=\gamma_1+\tfrac{\pi}{2}$ in the second observation. This can be summarized by the specific application of formula (\ref{equ:multiobserv_objfunc}):
\begin{align*}
\text{arg max}_{\,(\varphi,w)}\; \frac{1}{N} \sum\limits_{i=1}^{N} \iiint\limits_{\mathbb{R}^3} g_1\left(\left(\begin{array}{ccc}
\cos(\gamma_1) & \sin(\gamma_1) & 0\\
0 & 0 & 1
\end{array}\right)\cdot\vec{p}_1(\vec{z};(\varphi,w)) \right) \cdot\dots\\
 g_2\left(\left(\begin{array}{ccc}
\cos(\gamma_1+\frac{\pi}{2}) & \sin(\gamma_1+\frac{\pi}{2}) & 0\\
0 & 0 & 1
\end{array}\right)\cdot\vec{p}_1(\vec{z};(\varphi,w)) \right)\cdot q_i(\vec{z})\;\text{d}\vec{z}\;,
\end{align*}
and approximated by Monte--Carlo integration by the specific application of formula (\ref{equ:multiobserv_objfunc_MC}):
\begin{align*}
\text{arg max}_{\,(\varphi,w)}\; \frac{1}{N\, R} \sum\limits_{i=1}^{N} \sum\limits_{r=1}^R \;g_1\left(\left(\begin{array}{ccc}
\cos(\gamma_1) & \sin(\gamma_1) & 0\\
0 & 0 & 1
\end{array}\right)\cdot\vec{p}_1(\vec{z}_{i,r};(\varphi,w)) \right) \cdot\dots\\
 g_2\left(\left(\begin{array}{ccc}
\cos(\gamma_1+\frac{\pi}{2}) & \sin(\gamma_1+\frac{\pi}{2}) & 0\\
0 & 0 & 1
\end{array}\right)\cdot\vec{p}_1(\vec{z}_{i,r};(\varphi,w)) \right) \;,
\end{align*}
with $\vec{z}_{i,r}$ sampled from the densities $q_i(\vec{z})$.\smallskip

A short comment regarding the calibration of this camera system: It has to be noted that for the calibration of this specific geometry parameter $\gamma_1$ can be set arbitrarily, since the only relevant information is that $\gamma_2:=\gamma_1+\tfrac{\pi}{2}$. In practice one would (if there are no arguments against that) set $\gamma_1:=0$ for the calibration and let $\gamma_2$ optimize in order to find the deviations from the relative observation angle $\tfrac{\pi}{2}$.

\subsubsection{Lateration System with Feature Coordinate Densities and Rigid Affine Transformation Geometry}
\label{sec:laterationexample}

As second example, we are focusing on a ranging/lateration system, which measures the distances from a lateration point to each feature, such as radar or lidar devices. This naturally leads to $1$D observations of the $3$D objection consisting of the feature coordinates $\vec{z}_i$ ($i=1,..,N$). We assume to not know the position of the feature coordinates exactly, and, in consequence, utilize further unspecified density functions $q_i(\vec{z})$.

\begin{figure}[htbp]
\centering
\includegraphics[width=11cm]{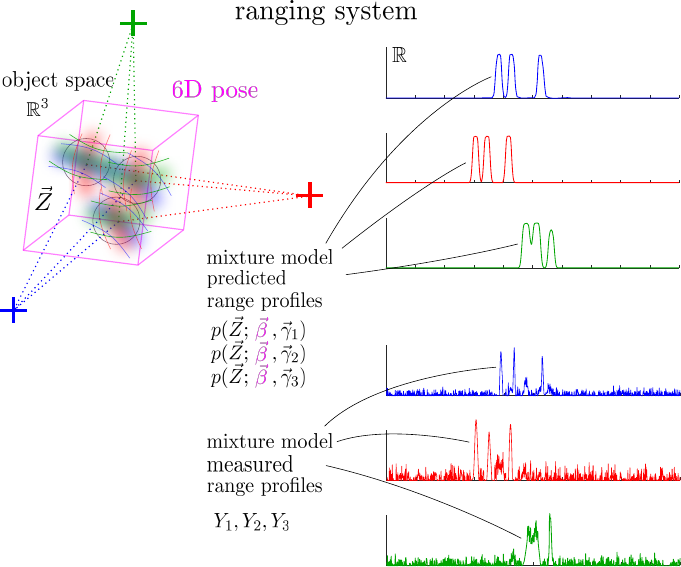}
\caption{Schematic illustration of a ranging system observation geometry and their connections to the mixture model approach.}
\label{OverviewPlot_B}
\end{figure}

The projection function in this application consists of two steps. First, the coordinate transformation of the object as the general \textit{rigid affine transformation}, and second, the projection to distances. We get with $\vec{\beta}:=(V,\vec{w})$ and $\vec{\gamma}$ the position of the lateration source in $3$D:
\begin{align*}
\vec{u} := \vec{p}_1\left(\vec{z};(V,\vec{w})\right) &:= V^T\cdot\left(\vec{z}+\vec{w}\right)\\
p_2(\vec{u};\vec{\gamma}) &:= \sqrt{\left(u_{1}-\gamma_1\right)^2+\left(u_{2}-\gamma_2\right)^2 + \left(u_{3}-\gamma_3\right)^2}\;.
\end{align*}
The overall measurement situation is demonstrated in figure \ref{OverviewPlot_B}.

With this example, it is demonstrated that the argument presented in this paper does neither depend on well--behaved pdfs in the modeling process (such as Gaussians) nor on a specific type of observation system. The objective function for one projection stays general and we arrive at the formula including the Monte--Carlo approximation as specific application of formula (\ref{equ:singleobserv_objfunc_MC}):
\begin{align*}
\text{arg max}_{\,(V,\vec{w})}\; \frac{1}{N\,R} \sum\limits_{i=1}^{N} \sum\limits_{r=1}^{R} g\left(p_2\left(V^T\cdot(\vec{z}_{i,r}+\vec{w}); \vec{\gamma}\right)\right)\;,
\end{align*}
subject to $V\in\mathbb{R}^{3\times 3}$ with orthonormal columns and with $\vec{z}_{i,r}$ sampled from the densities $q_i(\vec{z})$. It has to be mentioned that in this general formulation the specific orthogonality of $V$ directly leads to constraints in optimization, such as a system of nonlinear equations for the entries of $V$, which would need to be dealt with in the optimization routine. Of course, orthogonality of $V$ can be also enforced by the construction of the matrix, such as a rotation matrix and utilize the angles as degrees of freedom as demonstrated in the results section \ref{sec:accuracy_res}.  \smallskip

For the direct extension to $3$ projections we utilize three independent lateration devices, where $\vec{\gamma}_l$ ($l\in\{1,2,3\}$) are the three positions of the lateration devices. We get by the specific application of formula (\ref{equ:multiobserv_objfunc_MC}) utilizing Monte--Carlo integration:

\begin{align*}
\text{arg max}_{\,(V,\vec{w})}\; \frac{1}{N\,R} \sum\limits_{i=1}^{N} \sum\limits_{r=1}^{R} \prod\limits_{l=1}^3 g_l\left(p_2\left(V^T\cdot(\vec{z}_{i,r}+\vec{w}); \vec{\gamma}_l\right)\right)\;,
\end{align*}
subject to $V\in\mathbb{R}^{3\times 3}$ with orthonormal columns and with $\vec{z}_{i,r}$ sampled from the densities $q_i(\vec{z})$.\smallskip

A short comment regarding the calibration of this ranging/lateration system: The three lateration devices have specified coordinates $\vec{\gamma}_1,\vec{\gamma}_2, \vec{\gamma}_3$. Since the absolute position of the system in a global coordinate system is not important for practical measurements with that system, one can set $\vec{\gamma}_1 = (0,0,0)^T$, $\vec{\gamma}_2=(c_{2,1},0,0)^T$ and $\vec{\gamma}_3 = (c_{3,1},c_{3,2},0)^T$ and, consequently, obtain pose estimations relative to this defined geometry.

\section{Results}
\label{sec:results}

\subsection{Proposed Strategy for the Simulation}
\label{sec:res_proposed_strategy}

In this numerical results section the use of this unifying modeling framework is demonstrated by comparing three systems, which are directly related to the previous introduced application examples. Specifically, we are working with an orthogonal two camera system for systems 1 and 2, with the difference that system 1 uses exactly known feature positions of the model (utilizing delta distributions) and for system 2 Gaussian pdfs. System 3 is a lateration system with three lateration points utilizing the same pdfs for the feature positions as system 2. It is pointed out that the comparison of two cameras for system 1 and 2 and three lateration points for system 3 is fair since for both system types this is the lowest number of detectors necessary in order to define the position of a single point in 3D space.\smallskip

The strategy behind this is
\begin{itemize}
\item[a)] to investigate the use of different knowledge levels of the feature coordinates in 3D space in the same camera measurement setup by comparing system 1 and 2, as well as,
\item[b)] to compare different observation setups / principles directly by comparing system 2 and 3.
\end{itemize}

\begin{figure}[htbp]
\centering
\includegraphics[width=12cm]{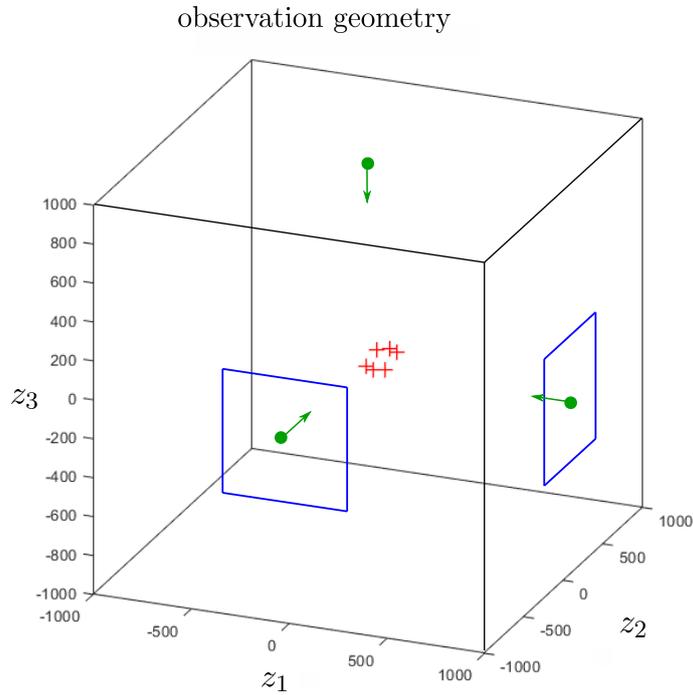}
\caption{Illustration of the two different observation geometries. Red Crosses: feature coordinates, Blue boxes: the two orthogonal projections of system 1 and 2, Green dots and arrows: three points of lateration sources for system 3.}
\label{comparison_geometry}
\end{figure}

Further, we are utilizing the coordinates of an object consisting of six observable feature points, which are presented in table \ref{tab:featurepoints}. This is an asymmetric rigid body, avoiding ambiguities under the given pose transformations. The observation geometries and the object coordinates are illustrated in figure \ref{comparison_geometry}. \smallskip
\begin{table}
\centering
\begin{tabular}{l||l|l|l|l|l|l}
feature point & 1 & 2 & 3 & 4 & 5 & 6 \\\hline\hline
$x$&             0     &   100.00      &       0    &    100.00   &     100.00   &          0\\
$y$&        -25.00     &   -25.00      &   61.60    &    111.60   &      25.00   &     111.60\\
$z$&        -43.30     &   -43.30      &  -93.30    &     -6.70   &      43.30   &      -6.70
\end{tabular}
\caption{Example feature coordinates of the object utilized in the demonstration in this results section.}
\label{tab:featurepoints}	
\end{table}

In the proposed simulation strategy, we will present four simulation scenarios $I$--$IV$ for the three described systems (leading to $12$ simulation setups), which investigates the use of the stochastic--geometrical modeling framework under specific challenging situations, which are of special interest in practice:
\begin{itemize}
\item[$I$)] The base scenario with normally distributed feature point density functions $q_i(\vec{z})$ in object space with standard deviation $\sigma = 10$ in subsection \ref{sec:res_vis} and $\sigma = 5$ in subsection \ref{sec:accuracy_res}. Further the error / sensing density function $\varepsilon(\vec{y})$ is also normally distributed with $\sigma_{\varepsilon}=5$. Simulation scenarios $II$--$IV$ are deviations from this base scenario.
\item[$II$)] In this scenario, the standard deviation of the error / sensing density function is strongly increased by $\sigma_{\varepsilon}=20$, which corresponds to a very low resolution power of the measurements / feature extraction. We will investigate the consequence of such low quality measurements.
\item[$III$)] In this scenario, the effect of a rigid body deformation of the object is investigated. The deformation of the object is applied directly before observation without updating the object model leading to an object model error in the estimation. The object deformation is realized by a coordinate--dependent scaling around the geometrical center of feature points: all $z_1$--coordinates are multiplied by $0.8$, the $z_2$--coordinates are kept as is and the $z_3$--coordinates are multiplied by $1.2$.
\item[$IV$)] In this scenario, the dependency of the objective function on extreme measurement noise which is at the level of SNR $\approx 0.5$ with a Gaussian noise function is demonstrated. Signal--to--Noise--Ratio (SNR) is defined in this context by maximum signal of one feature appearance divided by the standard deviation of the Gaussian. This noise level is applied to both camera signals as well as the lateration system signals.
\end{itemize} 

These $12$ simulation setups are investigated in two ways: In subsection \ref{sec:res_vis} we will perform a qualitative comparison by presenting the objective functions for a reduced $2$D pose estimation visually, and in subsection \ref{sec:accuracy_res} we will focus on a quantitative comparison of a full $6$D pose estimation.

\subsection{Visualization of the Objective Function}
\label{sec:res_vis}

In order to achieve a reasonable visual comparison, we utilize the described object with the same rigid body transformation. The center of the visual comparison is the plot of the objective function, which is generated for all three systems under varying circumstances. In order to achieve a direct impression of the objective function, we utilize for all systems the diagonally shifted and around $z_3$--axis rotated transformation $\vec{p}_1(\vec{z};(\varphi,w))$ as presented in subsection \ref{sec:cameraexample}. For the first two systems the same (orthogonal parallel beam camera projection) $\vec{p}_2$ is utilized as described in the camera application subsection \ref{sec:cameraexample}. For the third system (lateration) we utilize the observation function $p_2$ as described in lateration example application subsection \ref{sec:laterationexample}.\smallskip

\begin{figure}[htbp]
\centering
\includegraphics[width=14cm]{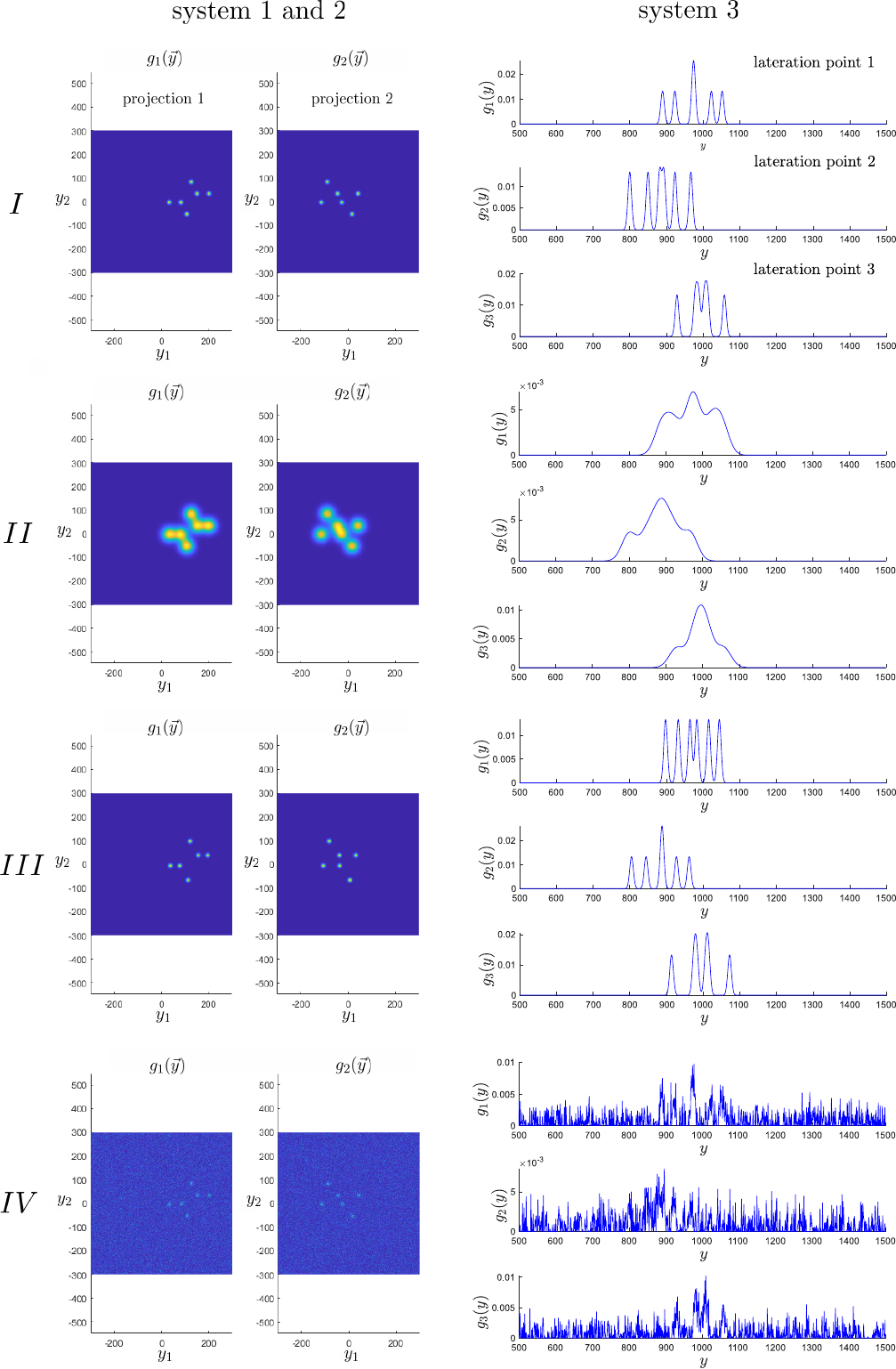}
\caption{Illustration of the two different simulated measurements types for various simulation scenarios. Left column: orthogonal camera observation measurements for system 1 and 2, Right column: Lateration observation measurements for three lateration points. First and second row: Presentation of different measurement error / sensing density standard deviations with $\sigma_{\varepsilon}=5$ ($I$) and $\sigma_{\varepsilon}=20$ ($II$). Third row ($III$): illustration of the observations with an object model error. Fourth row ($IV$): Simulation of strong noise levels with SNR $\approx 0.5$ for the camera as well as the lateration system applying a Gaussian noise function.}
\label{comparison_measurements}
\end{figure}

We simulate artificial measurements, first, by assuming a true value for $\vec{\beta}=(\varphi,w)=(\tfrac{\pi}{3},40)$ which we want to determine by the estimations, second, applying the described rigid body transformation $\vec{p}_1$ to the feature points and, third, performing the projection within the corresponding system. In consequence, the artificially generated measurements for system 1 and 2 are identical camera projections and for system 3, we get three individual depth profiles. The simulated projections of all scenarios $I-IV$ are presented in figure \ref{comparison_measurements}.\smallskip

\begin{figure}[htbp]
\centering
\includegraphics[width=14cm]{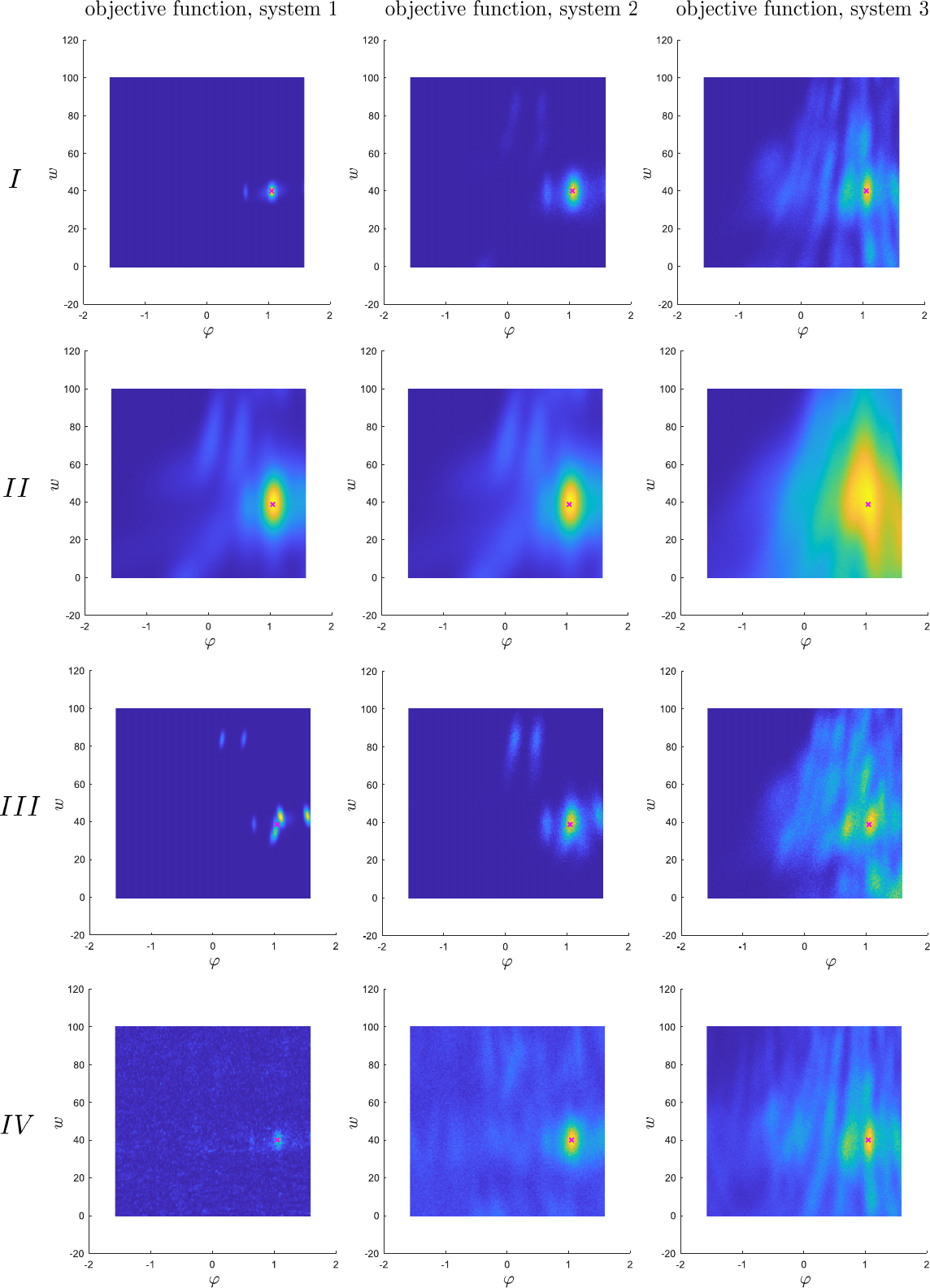}
\caption{Illustration of the different objective functions depending on the parameters $\varphi$ and $w$ (from left to right column) of system 1 and 2 (orthogonal camera observations) to system 3 (three point lateration). The true value of rigid body transformation parameters are presented as magenta cross. First and second row: Presentation of different measurement error / sensing density standard deviations with  $\sigma_{\varepsilon}=5$ ($I$) and $\sigma_{\varepsilon}=20$ ($II$). Third row ($III$): presentation of the effect of an object model error on the objective function without (system 1) or with (system 2 and 3) incorporated object model uncertainties. Fourth row ($IV$): illustration of the robustness of the objective function with respect to strong noise levels.}
\label{comparison_objfuction}
\end{figure}

The direct comparison of the resulting objective functions reflects the differences of the knowledge level about the point features (between system 1 and 2) and the geometric construction (between system 2 and 3) of the three observation systems. It is presented in figure \ref{comparison_objfuction}. It has to be noted that the objective function for system 2 and 3 was generated by the Monte--Carlo integration as presented in the general methods section with $R=240$ for system 2 and $R=120$ for system 3 for each pixel.\smallskip

First, the influence of the sensing density (scenario $II$) shows how the objective function blurs as the measurement resolution degrades compared to the base scenario $I$. An artificially blurred sensing can be utilized for the start of an optimization with a successive reduction of $\sigma_{\varepsilon}$. Reducing the standard deviation $\sigma_{\varepsilon}$ leads numerically also to the necessity to increase the Monte--Carlo sample size in order to achieve a lower noise level in the objective function. \smallskip

Second, the object model error by deformations (scenario $III$) leads to an important observation regarding the use of the random variable model for the object feature points. System 1 is not able to cope with object model error well, leading to multiple local maxima besides the true value in the objective function. This is due to the failing rigid body assumption of the feature coordinates. This effect is different for system 2 which still shows a distinct maximum at the correct pose values for $(\varphi,w)$ due to the direct incorporation of the object model uncertainties in the feature coordinates $\vec{Z}_i$, i.e. we are essentially not directly violating these more flexible assumptions. The lateration system 3 benefits also from the incorporated object model uncertainty by showing high values in the correct region of the objective function. Compared to system 2, system 3 shows a slightly shifted maximum leading to the assumption that the lateration system is slightly more sensitive to object model errors than the camera system in this application, which is an interesting result on its own. 

Third, due to the inherent averaging when calculating the objective function during integration for systems 2 and 3, the stochastic description of the problem proves to be robust against extreme measurement noise as demonstrated in case $IV$. It was not the purpose to present the most realistic noise characteristic but to show how robust the objective function behaves under extreme distortions. For system 1 the noise is much more visible in the objective function due to the reduced averaging.

\subsection{Accuracy Evaluation}
\label{sec:accuracy_res}

Besides the qualitative, visual evaluation, we also want to investigate the effectiveness and accuracy of the proposed method quantitatively for the three systems presented in the previous subsection \ref{sec:res_vis}. This works as a numerical validation and plausibility study of the stochastic--geometrical modeling and the presented algorithm. Of course, in this section we are not able to investigate all possible application scenarios for this type of modeling, but should demonstrate its utility by selected simulation scenarios.\smallskip

This time, we will demonstrate the full $6$D pose estimation. For this reason, we need to define the rigid affine transformation with the parameters $\vec{\beta}:=(\vec{\varphi},\vec{w})$ with the three rotation angles $\vec{\varphi}\in\mathbb{R}^3$ and translational components $\vec{w}\in\mathbb{R}^3$ by
\begin{align*}
\vec{p}_1(\vec{z};(\vec{\varphi},\vec{w})) := R(\vec{\varphi})\cdot (\vec{z} + \vec{w})\;,
\end{align*}
and the rotation matrix $R(\vec{\varphi}) = R_3(\varphi_3)\cdot R_2(\varphi_2)\cdot R_1(\varphi_1)$ with $R_i$ the rotation matrix around the $i$th axis.\smallskip

The two observational functions are
\begin{align*}
\vec{p}_2(\vec{u};\gamma_l) := \left(\begin{array}{ccc}
\cos(\gamma_l) & \sin(\gamma_l) & 0 \\ 0 & 0 & 1
\end{array}\right)\cdot \vec{u}
\end{align*}
for system 1 and 2 with $L=2$ and $\gamma_1=0$, $\gamma_2=\pi/2$, and
\begin{align*}
p_2(\vec{u};\vec{\gamma}_l) &:= \sqrt{\left(u_{1}-\gamma_{l,1}\right)^2+\left(u_{2}-\gamma_{l,2}\right)^2 + \left(u_{3}-\gamma_{l,3}\right)^2}\;.
\end{align*}
for system 3 with $L=3$ and $\vec{\gamma}_1 = [0,-1000,0]$, $\vec{\gamma}_2 = [1000,0,0]$ and $\vec{\gamma}_3 = [0,0,1000]$.\smallskip

For demonstration purposes, we will randomly sample the true values $(\vec{\varphi},\vec{w})$ for each simulation run from a uniform distribution on the intervals $\varphi_1,\varphi_2\in[-\pi/36,\pi/36]$ and $\varphi_3\in[-\pi/3-\pi/36,-\pi/3+\pi/36]$, as well as $w_1,w_2,w_3\in[35,45]$ ($\pi/36$ corresponds to $5^{\circ}$, which means in this simulation a $\pm 5^{\circ}$ variation of each true angular value is illustrated, which is regarded as a significant range relative to the expected accuracy). These intervals correspond to a similar setting as in subsection \ref{sec:res_vis} with additional variations of the presented objective functions in figure \ref{comparison_objfuction}, in order to provide a continuation to the previous results.\smallskip

For the realization of the optimization in each simulation run a standard routine is selected with the \textit{Nelder--Mead--}Algorithm of Matlab's \textit{fminsearch}--function. The starting values of this local optimization are selected also randomly by varying the true values $\vec{\varphi}+\Delta\vec{\varphi}$ and $\vec{w}+\Delta\vec{w}$ with an additional uniform sampling from the intervals $\Delta\varphi_1,\Delta\varphi_2,\Delta\varphi_3\in[-\pi/72,\pi/72]$ (i.e. $[-2.5^{\circ}, 2.5^{\circ}]$) and  $\Delta w_1,\Delta w_2,\Delta w_3\in[-5,5]$. The random selection of the true values for each simulation run as well as the random selection of starting values for the optimization has the goal to demonstrate the effectiveness and overall accuracy of the algorithm for the three different systems under varying situations. \smallskip

Regarding the computational effort, the number of samples in the Monte--Carlo integration is selected for system 2 and 3 to be constant with $R=1000$ and sampled by a normally distributed \textit{latin hypercube sampling} by Matlab's \textit{lhsnorm}--function. In total, for the implementation a standard tooling is utilized with only a moderate focus on a computational runtime. In order to judge the computational efficiency, the most runtime is due to interpolation of the observations $g_l$. For system 2 a 2D \textit{bilinear} interpolation routine is utilized for evaluating the two 2D images $g_1$ and $g_2$, and for system 3 a 1D \textit{linear} interpolation is utilized for evaluating the three 1D signals $g_1, g_2, g_3$. The runtime for one evaluation of the objective function (recall, with $R=1000$ interpolations per feature and observation) for system 2 is on average $10.8$ milliseconds and for system 3 on average $10.2$ milliseconds on a standard working laptop with computation on a $1.8$ Ghz single core. For one $6$D pose estimation with the \textit{Nelder--Mead--}Algorithm in the most demanding scenario $IV$  approximately $300-1200$ objective function evaluations are necessary (strongly depending on each simulation run), leading to computational times between $3-13$ seconds per pose estimation. It must be noticed that in this evaluation it is not the focus to present the most efficient implementation of the proposed optimization problem but to show the typical accuracy behavior in a simulation study depending on scenarios $I-IV$. In consequence, we expect several factors of acceleration in computational runtime with high efficiency programming, possibly allowing real--time tracking applications.\smallskip

The core of this study is to apply the scenarios $I$--$IV$ of subsection \ref{sec:res_proposed_strategy} to the three systems for a full $6$D pose estimation with $M=300$ simulation runs for each system and scenario. The statistical evaluation of the residual estimation errors between the true pose values $(\vec{\varphi},\vec{w})$ and the estimated values are shown as RMS values in table \ref{tab:accuracy_results}. From this table one can observe the following points:\smallskip

\begin{table}
\centering
{
\begin{tabular}{ll||l|l|l|l|l|l}
 &  & $\varphi_1$ & $\varphi_2$ & $\varphi_3$ & $w_1$ & $w_2$ & $w_3$\\\hline\hline
system 1:&$I$  & 0.51 & 0.53 &  0.18 &  0.344 &  0.286 &  1.271\\
&$II$ & 0.84 &  1.18 &  0.81 &  1.025 &  1.389 &  2.783 \\
&$III$ & 0.44 &  0.70 &  0.21 &  0.325 &  0.303 &  1.353 \\
&$IV$ & 1.47 &  1.47 & 1.32 &  2.591 &  2.695 &  2.532\\\hline
system 2:&$I$  & 0.004 &  0.004 & 0.002 &  0.045 &  0.047 &  0.044 \\
&$II$ & 0.72 &  1.24 &  0.85 &  1.073 &  1.456 &  2.924  \\
&$III$ & 0.003 &  0.004 &  0.002 &  0.047 &  0.044 &  0.047  \\
&$IV$ & 0.56 & 0.57 &  0.31 &  0.567 &  0.566 &  1.301\\\hline
system 3:&$I$  & 1.36 &  1.24 &  1.14 &  1.648 &  1.712 &  2.767\\
&$II$ & 11.53 &  11.25 &  10.62 & 11.184 & 16.771 & 13.312 \\
&$III$ & 0.67 &  1.13 & 0.85 &  1.141 &  1.364 &  2.426\\
&$IV$ & 1.73 &  1.61 & 1.26 &  2.028 &  1.978 &  3.280\\
\end{tabular}}

\caption{Presentation of the statistical evaluation of the full $6$D pose estimation results for system 1, 2 and 3, as well as for each simulation scenario $I$--$IV$. The residual errors are presented as RMS--values of all $6$D parameters with the angular units $\varphi_i$ in [deg] and $w_i$ in the length unit.}
\label{tab:accuracy_results}
\end{table}

First, system 2 outperforms system 1 in essentially all simulation scenarios $I$--$IV$ and demonstrates the benefit of the stochastic--geometrical argumentation which utilizes pdfs for the feature points in object space in system 2 in contrast to system 1, which does not account for feature position flexibility. It is remarkable that this benefit also holds true for the case of no strong distortions (scenario $I$), since the use of pdfs for the feature points introduces a problem--specific averaging. This is in accordance to the observation in the previous visualization of the objective function.\smallskip

Second, system 2 outperforms system 3 in essentially all simulation scenarios, which means that under the given parameter set the two orthogonal camera system is in general better than the orthogonal trilateration system. This is in accordance to the observations in subsection \ref{sec:res_vis}, which shows a less distinct maximum for system 3 than for system 2 in the objective function. It has to be noted that this does not mean that the trilateration system could not be better in practice, since the parameter of distance measurement accuracy $\sigma_{\varepsilon}$ could be much lower for lateration systems (based on laser interference) relative to camera systems with given pixel resolutions. It is pointed to the fact, that within this unifying modeling framework it is possible to determine how much the distance measurement accuracy in the lateration system must be better to outperform the camera setup.\smallskip

Third, scenario $II$ shows that for low--resolution measurements all systems 1--3 are affected strongly, even to the point of uselessness of the estimation. If the inherent resolution of the measurement equipment is too low for the $6$D pose estimation, a different placement of the measurement devices or an increased number of measurement devices could be investigated within this framework. A general tipping point for low--resolution measurements is that the individual point features cannot be separated anymore in the observation. This depends, of course, on the physical distance of the features and their arrangement on the object. Such a problematic situation is demonstrated in figure \ref{comparison_measurements} where the individual feature observations are not possible anymore in scenario $II$ for system 3, leading to the worst pose estimation results in this quantitative analysis.\smallskip

Fourth, scenarios $III$ and $IV$ demonstrate by comparing systems 1 and 2, that the estimation utilizing the stochastic geometric model is much more robust against strong object deformations ($III$) as well as strong noise levels of the measurement ($IV$). This is also in accordance to the observations of subsection \ref{sec:res_vis}.

\section{Discussion}

In this work it is demonstrated how object pose estimation can be modeled thoroughly by stochastic arguments utilizing mixture models. The author is not aware of such a presentation in the literature and, therefore, this approach is regarded as original and practically helpful for everyone modeling such observation processes. A major challenge of such an applied presentation is to show the general argument on the one side, and to present expressive examples that help understanding the impact on the other side. This is resolved by, first, presenting the general argumentation and, second, focusing on classical camera and lateration systems, which are presented and compared within this framework. Especially the numerical results show how different knowledge levels or measurement device setups can be compared within this unifying modeling framework. An insightful comparison is demonstrated by a qualitative presentation of the objective functions of each selected setup and simulation scenario since the difficulty of estimation is directly reflected by the objective function. In addition, a numerical evaluation of full $6$D pose estimation over all simulation scenarios is presented which shows how to quantify the estimation accuracy. Overall, the benefit of utilizing the mixture model pdfs is demonstrated by the numerical simulations since they show {the possibility of} high accuracy and robustness while simultaneously avoiding the explicit solution of the correspondence problem. It is obvious that utilizing this framework many more measurement device setups/technologies and measurement scenarios can be compared than the presented examples.\smallskip

When constructing/designing an observation system practically, a major challenge is how to find suitable geometries and/or detector types. It can be argued that such a unified framework allows finding such a system setup, even when different observation setups or measurement principles are compared or mixed, since a fair comparison can be achieved within this framework. This \textit{inter--technological} perspective on the mathematical modeling seems unusual in the presentation of observation processes. Further, this framework also shows the general approach of how to calibrate such systems as it is explained in the text. Certainly, this general perspective is crucial for a sound argumentation for or against certain system designs for the application at hand.\smallskip

From a practitioner's perspective, defining the necessary density functions in this new framework might be challenging since each of them should be backed up by independent observations, e.g. the feature extraction error density or realistic feature position densities in object space. However, this effort has its own merit, since this new type of blurry information is transparently incorporated into the estimation process applying this framework (which benefit was presented in the numerical results section). If these independent pieces of information are not available, reasonable assumptions must be defined  such as working with specification sheets of system components (e.g. point spread functions of a camera lens) or typically expected uncertainties in comparable observation scenarios. In such situations, it is in general difficult to assess the influence of modeling errors since it is highly application dependent. Since this approach is focusing on a stochastic best fit, minor modeling errors in the densities are expected to have only minor influence on the pose estimation. 
It is also important to notice, that finding advantageous geometric parameter sets $\vec{\gamma}$ in the observation model might be challenging, especially, if one wants to define which parameters are set to fixed values and which are optimized in a calibration routine of a multi--view (and possibly multi--technological) system.\smallskip

In this presentation certain assumptions related to the feature selection are used that might lead to practical challenges before being able to apply this framework. First, it is not resolved how the object model is obtained but assumed that there exists a suitable object model. A straightforward approach in industrial applications is the use of a CAD of the object model. Second, it is not discussed how the feature points are selected at the object model, especially with the focus of visibility in the observations. This can be realized by texture or object shape, hand selected or automated, and certainly depends on the observation technology (e.g. cameras including illumination of the scene vs. ranging systems). Third, it certainly is an effort preparing the real observation measurement to get the feature appearance mixture model density with a reasonable quality. This might involve major challenges applying adaptive image processing and segmentation algorithms, which certainly will depend on each application such as the observation device and feature type. These points are regarded as independent of this framework presentation, since there is a plurality of established methods, which are applied for feature extraction in camera and lateration systems.\smallskip

In this paper we propose a model driven first principles approach by defining random variables and show how they are transformed in an observation scenario to estimate an object pose. We acknowledge that this in the contrast to the algorithmically driven current main body of research, which is based on (deep learning) artificial neural networks architectures. We want to stress that these papers are very successful and mainly rely on an extensive training for specific camera setups and exemplary video stream data which is shared in the computer vision community (e.g. Kendall \textit{et al.}\cite{Kendall2015}) and Rad \textit{et al.}\cite{Rad2017}). Compared to this, the goal of this paper is to allow to design and compare new observation scenarios and devices. In consequence, we do not think that it is reasonable to compare the general argumentation framework resulting in a general algorithmic family (with plausibility simulations in the results section) of this presentation to specifically highly trained algorithmical approaches on example video stream data sets which are generated by given camera devices. On the one hand, we do encourage future work in developing and adapting the presented algorithmic family further to specific applications so that comparison to existing algorithmical approaches can be reasonably performed. On the other hand, we are optimistic that combining the probabilistic view of this modeling framework with algorithm driven solutions can lead to improvements of pose estimation in the future.\smallskip

This framework is presented for the practically relevant application of finding the pose of a 3D rigid object, which has observable point features. The formulation of the general mathematical arguments is done in an inclusive fashion in order to allow a more general/abstract application. Quite straightforwardly this might include, e.g. parameter estimation of nonlinear object transformations $\vec{p}_1$, geometric observation in more abstract high--dimensional feature spaces (not only real world object representations) or the extension of the meaning of the density $q(\vec{z})$ not only observing point features but larger observable structures with spatial extension. Since such extensions do not change the general mathematical argument but only the interpretation of components of this framework, it is regarded as a benefit of this description and thinking about further applications is encouraged.

\section{Conclusion}

In this paper a stochastic--geometrical modeling framework is introduced which can be applied for many object pose estimation approaches based on different observational techniques. Basis of this approach is the thorough stochastic interpretation of the measurement and modeling of the observation process, starting from feature coordinates, which are introduced as random variables with mixture model densities. This eventually leads to the avoidance of the correspondence problem between features in the observations and the object model. Characteristically different application examples are presented with a two orthogonal camera system and a three point lateration system. The unified handling of these different systems in one framework allows the direct comparison of these systems as demonstrated in the results section in order to gain deeper insight into the pose estimation problem and its technical realizations.


\end{document}